\documentclass[sigconf]{acmart}

\AtBeginDocument{%
  }

\setcopyright{acmlicensed}
\copyrightyear{2018}
\acmYear{2018}
\acmDOI{XXXXXXX.XXXXXXX}
\acmConference[KDD’25 SciSoc LLM Workshop]{KDD’25 SciSoc LLM Workshop: Large Language Models for Scientific and Societal Advances}{August 4, 2025}{Toronto, ON, Canada}

\usepackage{microtype}
\usepackage{hyperref}
\usepackage{subfiles}
\usepackage{xspace}
\usepackage{multirow}
\usepackage{multicol}
\usepackage{url}
\usepackage{booktabs}
\usepackage{longtable}
\usepackage{graphicx}
\usepackage{caption}
\usepackage{adjustbox}

\usepackage{lineno}

\definecolor{darkblue}{rgb}{0, 0, 0.5}
\hypersetup{colorlinks=true, citecolor=darkblue, linkcolor=darkblue, urlcolor=darkblue}

\newcommand{\benchname}{\textsc{CCSBench}\xspace}
\begin{document}

\title{\benchname: Evaluating Compositional Controllability in LLMs for Scientific Document Summarization}

\author{Yixi Ding}
\affiliation{
  \institution{National University of Singapore}
  \country{}
  }
\email{yixi.d@comp.nus.edu.sg}
\author{Jiaying Wu}
\authornote{Corresponding authors.}
\affiliation{
  \institution{National University of Singapore}
  \country{}
  }
\email{jiayingw@nus.edu.sg}
\author{Tongyao Zhu}
\affiliation{
  \institution{National University of Singapore}
  \country{}
  }
\email{tongyao.zhu@u.nus.edu}
\author{Yanxia Qin}
\affiliation{
  \institution{Singapore University of Technology and Design}
  \country{}
  }
\email{yanxia_qin@sutd.edu.sg}
\author{Qian Liu}
\affiliation{
  \institution{Sea AI Lab}
  \country{}
  }
\email{liuqian@sea.com}
\author{Min-Yen Kan}
\authornotemark[1]
\affiliation{
  \institution{National University of Singapore}
  \country{}
  }
\email{kanmy@comp.nus.edu.sg}

\renewcommand{\shortauthors}{Ding et al.}

\begin{abstract}
To broaden the dissemination of scientific knowledge to diverse audiences, it is desirable for scientific document summarization systems to simultaneously control multiple attributes such as length and empirical focus. However, existing research typically focuses on controlling single attributes, leaving the \textit{compositional control} of multiple attributes underexplored. To address this gap, we introduce \benchname, the first evaluation benchmark for compositional controllable summarization in the scientific domain. Our benchmark enables fine-grained control over both \textit{explicit} attributes (e.g., length), which are objective and straightforward, and \textit{implicit} attributes (e.g., conceptual or empirical focus), which are more subjective and abstract. We conduct extensive experiments using various large language models (LLMs) under various settings, including in-context learning, parameter-efficient fine-tuning, and two-stage modular methods for balancing control over different attributes. Our findings reveal significant limitations in LLMs capabilities in balancing trade-offs between control attributes, especially implicit ones that require deeper understanding and abstract reasoning. 
\footnote{Data is available at: \url{https://huggingface.co/datasets/dyxohjl666/CCSBench}.}\end{abstract}

\maketitle

\section{Introduction}
\label{sec:intro}

\begin{figure}[t]
  \centering
  \includegraphics[width=\columnwidth]{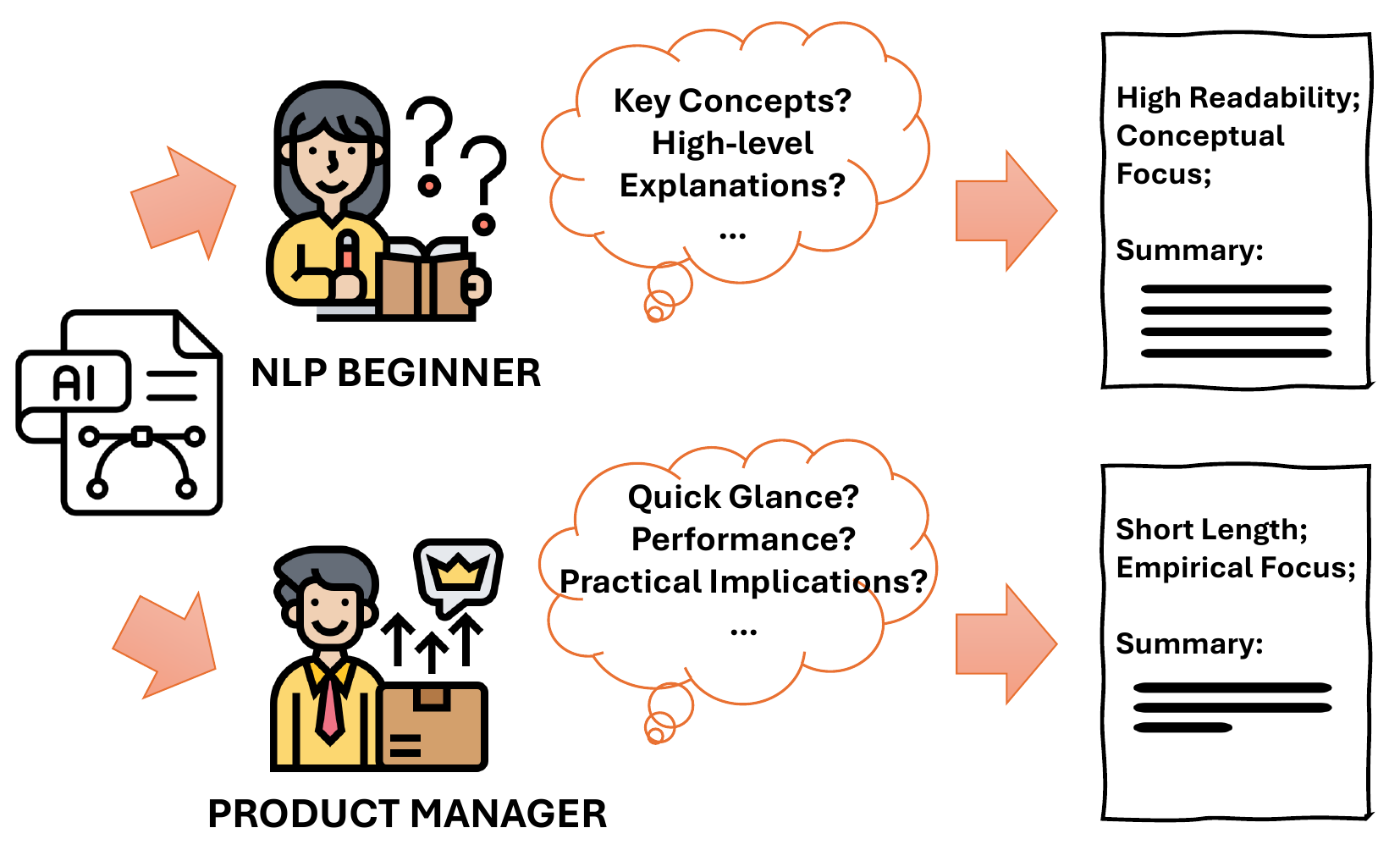}
      \caption{Compositional controllability allows scientific summaries to be tailored to diverse reader needs, addressing specific reader requirements.}
  \label{fig:userneeds}
\end{figure}

\textit{Compositional controllability} -- the ability to generate tailored summaries based on multiple attributes -- is essential for scientific document summarization systems to effectively communicate research to diverse audiences. Scientific papers contain complex information, but different readers have distinct needs and preferences when engaging with this content. For instance, as illustrated in Figure~\ref{fig:userneeds}, an NLP beginner may seek a clear, concept-focused explanation, while a product manager may prioritize a concise summary highlighting the model’s empirical performance. By dynamically adjusting summaries to align with varying expertise levels and interests, compositional control ensures that scientific knowledge is more accessible, relevant, and engaging for a broader audience.

Despite the broad demand for compositional controllability in scientific summarization, this capability remains largely underexplored in controllable text generation research. Most current efforts focus on using large language models (LLMs) to generate non-academic texts such as movie reviews~\cite{liu-etal-2024-multi} and news articles~\cite{chen2024can}, which are less complex than scientific documents adhering to strict academic conventions. Preliminary investigations into compositional scientific summarization~\cite{ding-etal-2023-cocoscisum} address only straightforward control requirements such as length and keyword inclusion, overlooking more subtle user requirements such as focus on empirical aspects. Developing methods for real-world deployment requires benchmarks that not only incorporate more sophisticated control attributes but also capture their interplay in a truly compositional manner, a challenge that remains unresolved.

In this paper, we introduce \benchname,the first evaluation benchmark for compositional scientific summarization that enables fine-grained control over multiple attributes. Drawing inspiration from Kahneman's cognitive theory~\cite{kahneman2013thinking}, which distinguishes between the fast, intuitive ``System 1'' and the slow, deliberate ``System 2'' modes of thinking, \benchname incorporates both \textbf{explicit} and \textbf{implicit} control attributes. Explicit attributes (e.g., length) are objective, easily quantifiable, and straightforward for both LLMs and humans to process. In contrast, implicit attributes (e.g.,  empirical focus) require deeper reasoning and human-like understanding, making them more challenging for LLMs to control effectively.

\begin{figure*}[htbp]
  \centering
  \includegraphics[width=17cm]{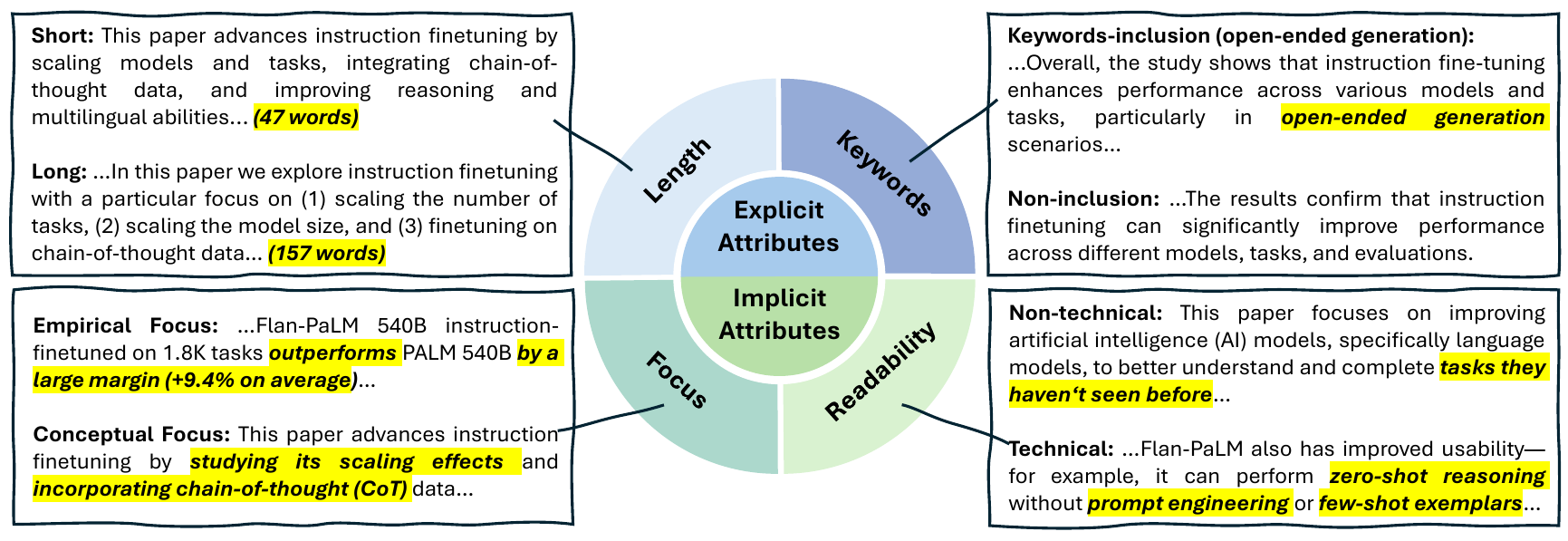}
      \caption{Illustration of explicit and implicit control attributes in \benchname.}
  \label{fig:cococtrl}
\end{figure*}

As illustrated in Figure~\ref{fig:cococtrl}, \benchname is structured around two explicit attributes: (1) \textit{length} and (2) \textit{keyword inclusion}, and two implicit attributes: (3) \textit{readability} and (4) \textit{focus}. \textit{Length} sets the target word count for the summary, while \textit{keyword inclusion} ensures the presence of specific terms in the output. \textit{Readability} controls language complexity and accessibility, tailoring the summary for technical or general audiences. \textit{Focus}, short for ``Empirical Focus Level'', adjusts the emphasis on empirical evidence versus conceptual aspects, requiring deeper contextual understanding and abstract reasoning. By dynamically adjusting these attributes, we curate diverse scientific summaries -- such as concise, highly readable conceptual explanations -- that meet real-world needs and cater to diverse reader expectations.

We evaluate a diverse set of LLMs on \benchname, including both proprietary LLMs and fine-tuned open-source variants.  Results reveal significant limitations in managing trade-offs between control attributes, particularly due to inadequate abstract reasoning for implicit attributes. In parameter-efficient fine-tuning (PEFT), decoder-only models like LLaMA2~\citep{touvron_2023_llama2openfoundation} and Mistral~\citep{jiang2023mistral7b} struggle with long-term dependencies, while encoder-decoder models like Flan-T5~\citep{flan-t5} adapt more effectively. Further analysis shows that standard task composition methods, such as AdapterFusion~\cite{pfeiffer-etal-2021-adapterfusion}, are unsuitable due to oversimplified adapter aggregation. Our findings suggest the need for more targeted research on compositionality in scientific document summarization.

\section{Related Work} \label{sec:related}

\subsection{Controllable Summarization}

Controllable summarization has attracted significant research interest, with efforts focused on guiding summaries using attributes such as length, topic, keywords, and sentiment. Early work explored the impact of oracle-provided attributes on summary quality~\citep{fan-etal-2018-controllable, see2017pointsummarizationpointergeneratornetworks}. More recent studies have aimed at controlling individual attributes like length or keywords~\citep{chan-etal-2021-controllable-markov, he2020ctrlsumgenericcontrollabletext}, but these methods often treat attributes independently, limiting their scalability and effectiveness in multi-attribute scenarios. MacSum~\citep{zhang_2023_macsum} represents a step forward by supporting multi-attribute control in the news and dialogue domains, where the content is relatively less technical. In contrast, scientific texts pose unique challenges for controllable summarization due to their high degree of linguistic precision and domain-specific complexity. CocoScisum~\citep{ding-etal-2023-cocoscisum} extends control to both length and keywords within scientific papers but remains constrained to explicitly defined attributes. Our work, \benchname, advances the field by enabling fine-grained control over both explicit and implicit attributes, offering a broader and more nuanced evaluation framework tailored to scientific summarization.

\subsection{LLMs for Scientific Document Processing}

Large Language Models (LLMs) have demonstrated remarkable capabilities across a wide range of scientific document processing tasks, including idea generation~\citep{si2025can}, literature review synthesis~\citep{2024autosurvey}, research critique~\citep{faizullah2024limgenprobingllmsgenerating}, and the production of scientific news articles~\citep{pu-etal-2024-scinews-scholarly}. These models also show promise in controllable text generation~\citep{urlana2024controllable, zhou-etal-2024-evaluating, yun2025ultragen}, though existing work primarily evaluates them under single-constraint settings (e.g., adjusting formality). Despite the growing need to tailor scientific content for diverse audiences, the ability of LLMs to perform compositional controllable generation -- handling multiple constraints simultaneously -- remains underexplored. To address this gap, we introduce the \benchname\ benchmark, designed to systematically evaluate LLM performance under multi-attribute control. Our findings highlight the limitations of current models, particularly in handling implicit, conceptual attributes such as readability and aspect focus.

\section{\benchname Data Curation}
\label{sec:benchmark}

\subsection{Task Formulation}
We formally introduce the task and notation used throughout the paper. Each example consists of a scientific document $D$ accompanied by a set of control signals $\mathcal{C} = \{len, kw, focus, read\}$, where $len$ refers to length, $kw$ to keywords, $focus$ to empirical focus level, and $read$ to readability. The objective is to generate a summary $S$ that adheres to the constraints imposed by $\mathcal{C}$ while maintaining narrative coherence and preserving key information from the original document. 

We define the four control attributes for \benchname as follows: 

\textbf{Length} specifies the desired number of words in the summary, ranging from concise to detailed, and catering to different audience needs based on their familiarity with the topic or time constraints. Length is divided into five bins, each representing a 50-word range (e.g., Bin 0: 0-50 words; Bin 1: 51-100 words).

\textbf{Keywords} ensures the inclusion of key terms from the source document, enabling readers to quickly grasp the core elements and assess the paper's relevance. Keywords can be left empty or set as pre-defined terms to be included in the summary.

\textbf{Readability} adjusts syntactic complexity and vocabulary to match audience proficiency, making summaries more accessible or technical. This enhances comprehension and engagement. It has two levels: [normal, high].

\textbf{Empirical Focus Level (Focus)} adjusts emphasis between empirical results and theoretical contributions, with two levels: [low, high]. A high empirical focus highlights data-driven aspects like experiments and results, while a low empirical focus emphasizes theories, frameworks, and broader implications. This control tailors summaries to readers seeking practical insights or conceptual understanding.

\subsection{Dataset}
\label{sec:data}

We construct \benchname based on the arXiv dataset~\cite{clement2019usearxivdataset}, a comprehensive collection of scientific documents from the widely used online preprint repository, arXiv. \benchname focuses on papers in Computer Science and Artificial Intelligence. To ensure high-quality input with minimal noise, we use the introduction and conclusion sections (\textbf{I+C}) as the input text, a practice validated by prior research \cite{cachola-etal-2020-tldr,meng-etal-2021-bringing-facet} and our empirical analysis in Section \ref{sec:input}. Following established practices in scientific summarization research~\cite{cohan-etal-2018-discourse}, we use abstracts as reference summaries, leveraging their widely recognized factuality and comprehensiveness~\cite{stanfordtips}.

The benchmark comprises four single-attribute control datasets and one compositional control dataset, each split into training, validation, and test sets using a 60/20/20 ratio. Dataset statistics are summarized in Table~\ref{tab:dataset}.

\begin{table}[t]
    \centering
    \caption{\label{tab:dataset} \benchname statistics.}
    \begin{tabular}{l|ccc}
    \hline
    \multirow{2}*{\textbf{Dataset}} & \multicolumn{3}{c}{\bf{\#Samples / \#Docs}} \\
    \cline{2-4}
      & {\bf Train} & {\bf Val} & {\bf Test}  \\
    \hline
    Length & 2,400 / 1,561 & 800 / 705 & 800 / 698 \\
    Keywords & 2,029 / 2,029 & 677 / 677 & 677 / 677 \\
    Readability & 2,400 / 1,687 & 800 / 724 & 800 / 728 \\
    Focus & 2,332 / 1,659 & 779 / 709 & 768 / 703 \\
    Compositional & 2,400 / 1,590 & 800 / 655 & 758 / 364\\

    \hline
    \end{tabular}
\end{table}
\begin{table*}[t] 
    \centering
    \caption{\label{tab:humaneval} Evaluation results validate the high quality of LLM-generated summaries in \benchname.}
    \begin{tabular}{l|cccccc}
    \hline
    \textbf{Dataset} & FAC & FLU & CTRL & ROUGE-1 & ROUGE-2 & ROUGE-L \\
    \hline
    Readability & 0.98 & 0.93 & 0.92 & 53.88 & 21.81 & 41.31\\
    Focus & 0.91 & 0.93 & 0.93 & 51.25 & 34.23 & 41.41 \\
    \hline
    \end{tabular}
\end{table*}

\subsection{Dataset Construction}
\label{sec:ds-construction}
The construction of \benchname involves two main steps: (1) construction of single-attribute control datasets, followed by (2) the creation of a compositional control dataset.

\paragraph{Single Control Dataset} 

We begin with KWX~\cite{fu_2023_kwx}, a dataset that provides keywords for each scientific document sourced from arXiv. From this, we select 3,700 papers in the subjects of Computer Science and Artificial Intelligence to form the foundation of our controllable summarization dataset, using their abstracts as natural reference summaries. 

The \textbf{Keywords} dataset, $\mathcal{D}_{kw}$, is created by filtering out any keywords from the KWX dataset that do not appear in the abstracts, ensuring high-quality keywords.

To address the challenge of automatically assessing readability and focus, we use GPT-4~\citep{openai2024gpt4technicalreport} to assist in generating summaries related to these two control attributes. We sample 2,000 papers from our data source, and assign the original abstracts with ``normal readability''. GPT-4 rewrites each abstract into a more layman-friendly version while maintaining the original meaning. These rewritten abstracts are labeled as ``high readability''. Combining these two types of abstracts, we form the \textbf{Readability} dataset, $\mathcal{D}_{read}$.

For focus control, we task GPT-4 with identifying the conceptual and empirical components of each abstract, then rewriting them separately without altering their meaning. Summaries derived from conceptual components are labeled as having ``low empirical focus,'' while those from empirical components are labeled as having ``high empirical focus''. These summaries form the \textbf{Focus} dataset, $\mathcal{D}_{focus}$.

To construct the \textbf{Length} dataset, $\mathcal{D}_{len}$, we sample 4,000 instances from $\mathcal{D}_{Read}$ and $D_{focus}$. Word counts are rounded to the nearest 50-word bin to create a length-controllable dataset.

\paragraph{Compositional Control Dataset}
To manage the complexity of controlling all four attributes simultaneously, we construct a compositional control dataset using $\mathcal{D}_{read}$ and $\mathcal{D}_{focus}$, denoted as $\mathcal{D}_{CC}$. Since readability and empirical focus are already annotated, and keywords and length can be easily derived as described earlier, each instance in this dataset is tagged with 2-3 attributes. This allows us to create both training and validation sets with compositional control over multiple attributes.

While it is impractical to collect instances covering every attribute combination in the training set, we ensure comprehensive evaluation by constructing a test set that includes all four attributes. We sample 400 instances from $D_{focus}$, where the length, keywords, and empirical focus levels are pre-determined. Following the method used to generate $D_{read}$, we use GPT-4 to rewrite each summary for high readability, ensuring accessibility for a layman audience.

\subsection{Data Quality Validation}
\label{sec:data_quality}

We assess the quality of GPT-4-generated summaries in \benchname through automated and human evaluations. First, we compute ROUGE scores to assess fluency and content accuracy by comparing generated summaries with their original counterparts. Next, we conduct a human evaluation to further verify the summaries’ controllability and overall quality.

We sample 120 documents each from $D_{read}$ and $D_{focus}$ and recruit Amazon Mechanical Turk\footnote{\url{https://www.mturk.com/}} annotators for binary assessments on three aspects: (1) \textbf{Factual Consistency (FAC)}: whether the summary accurately reflects key points, (2) \textbf{Fluency (FLU)}: whether it is natural and well-written, and (3) \textbf{Controllability (CTRL)}: for Readability, whether the summary is easier to read than the reference; for Focus, whether empirically-focused summaries emphasize data collection, experimental setup, and results over conceptual aspects (details in Appendix~\ref{sec:human-eval}).

Table~\ref{tab:humaneval} shows high accuracy scores (>0.9) across all metrics, confirming the quality of \benchname’s summaries. Rigorous annotator selection and strong inter-annotator agreement (>0.85, Section~\ref{sec:mturk}) further validate these results. 

\subsection{Reliability of Human Evaluation}
\label{sec:mturk}
We implement rigorous measures to ensure the reliability of human evaluations. Annotators on Amazon Mechanical Turk are selected based on strict quality criteria, including a $\geq98\%$ HIT approval rate and a minimum of 500 completed HITs, ensuring experienced and reliable contributors.

To validate annotation reliability, we conduct a pilot study with 20 randomly sampled instances for readability- and focus-controlled summaries. Each summary is evaluated by two independent annotators, and inter-annotator agreement is calculated. As shown in Table~\ref{tab:iaa}, all agreement scores exceed 0.85, confirming evaluation consistency. Based on these reliable pilot results, we scale the evaluation to two larger sets of 120 randomly sampled instances, each assessed by a single annotator, as shown in Section \ref{sec:data_quality}.

\begin{table}[htbp] 
    \centering
    \caption{Inter-annotator agreement scores for assessment questions on readability (Appendix \ref{sec:read_eval}) and empirical focus (Appendix \ref{sec:focus_eval}).}
    \begin{tabular}{lccccc}
    \hline
        & Q1 & Q2 & Q3 & Q4 & Q5\\
    \hline
    Readability & 1.00 & 0.85 & 0.95  &0.90 & - \\
    Focus & 0.95 & 1.00 & 0.85 & 0.90 & 0.95 \\
    \hline
    \end{tabular}

    \label{tab:iaa}
\end{table}

\section{Experimental Setup}
\label{sec:experiments}
\subsection{Models}
We evaluate a range of closed-source and open-source models for compositional controllable summarization, including GPT-3.5 \cite{openai2022chatgpt}, GPT-4~\citep{openai2024gpt4technicalreport}, GPT-4o \cite{hurst2024gpt}, LLaMA2~\citep{touvron_2023_llama2openfoundation}, Mistral~\citep{jiang2023mistral7b}, and Flan-T5~\citep{flan-t5}. For closed-source models, we conduct few-shot experiments by randomly selecting three demonstrations per test sample. For open-source models, we explore parameter-efficient fine-tuning (PEFT) using Low-Rank Adaptation (LoRA)\cite{hu2021loralowrankadaptationlarge} to adapt them to our dataset. Additionally, we revisit controllable summarization baselines from the news domain, including hard prompt (HP)\cite{zhang_2023_macsum} and soft prefix tuning (SP)\cite{li-liang-2021-prefix}. Further implementation details, including training setups, are provided in Appendix \ref{app:experiments}.

\subsection{Evaluation Metrics}
\label{sec:eval_metrics}
We evaluate all models based on both summarization quality and attribute controllability. We employ ROUGE~\cite{lin-2004-rouge} to evaluate the overall quality. To assess attribute controllability, we employ the following metrics tailored to each attribute.

\paragraph{Length Control} We measure length controllability using the Mean Absolute Deviation (MAD) ~\citep{liu-etal-2018-length-ctrl-cnn}, and the Pearson Correlation Coefficient (PCC) ~\citep{liu-etal-2018-length-ctrl-cnn}, to evaluate the distance between the target and generated lengths.

\paragraph{Keywords Control} Keyword control is evaluated using the Success Rate (SR) ~\citep{he2020ctrlsumgenericcontrollabletext}, which measures the fraction of specified keywords present in the generated summaries through exact matching after stemming.

\paragraph{Readability Control} For readability, we calculate the Flesch-Kincaid Grade Level (FKGL) ~\citep{Kincaid1975DerivationON-fkgl}, where a lower score indicates higher readability. The difference in FKGL scores ($\delta_{FKGL}$) between the high readability setting ($F_{high}$) and the normal readability setting ($F_{normal}$) reflects the model’s ability to control this attribute. A higher $\delta_{FKGL}$ is desirable.

\paragraph{Empirical Focus Control} For empirical focus, we use GPT-4 to classify summaries into high or low empirical focus levels. Validated with 93\% accuracy on a manually labeled dataset, GPT-4 predictions are used to compute the F1 score for each category, assessing the model’s focus control. Prompt details are provided in Appendix~\ref{sec:prompt}.

\begin{table*}[htbp] 
    \centering
    \caption{Performance comparison of representative language models on \benchname, evaluating overall quality and four control attributes. \textbf{Arch} refers to model architectures, with \textbf{ED} for encoder-decoder and \textbf{DO} for decoder-only. \textbf{Bold} numbers indicate the best performance, while \underline{underlined} numbers represent the second-best.}
    \begin{tabular}{llccccccccc}
    \hline
    & \multirow{3}*{\bf Method} & \multirow{3}*{\bf Arch}& \multirow{3}*{\bf Quality} & \multicolumn{3}{c}{\bf Explicit Attributes} & \multicolumn{4}{c}{\textbf{Implicit Attributes}} \\
    \cmidrule(r){5-7} \cmidrule(r){8-11}
    &  &  &  &  \multicolumn{2}{c}{\bf Length} & {\bf Keywords} & \multicolumn{3}{c}{\bf Readability} & {\bf Focus} \\
    \cmidrule(r){4-4} \cmidrule(r){5-6} \cmidrule(r){7-7} \cmidrule(r){8-10} \cmidrule(r){11-11}
     & & & {ROUGE-L $\uparrow$} & {PCC $\uparrow$} & {MAD $\downarrow$} & {SR $\uparrow$} & {F$_{normal}$} & {F$_{high}$$\downarrow$} & {$\delta_{FKGL}$ $\uparrow$} & {F1 $\uparrow$} \\
    \hline
    \hline
    \multicolumn{10}{l}{\bf Zero-Shot LLMs}\\
    \hline
    A1 &Flan-T5-XL & ED  & 13.49  & 0.00 & 1.43 & 0.44 & 13.94 & 13.57 & 0.37 &  0.36\\
    A2 &Flan-T5-XXL & ED & 14.25 & 0.00 & 1.39 & 0.46 & 12.66 & 12.68 & -0.02 & 0.39 \\
    A3 & LLaMA2-7B &  DO & 17.10  & 0.23 & 1.11 & 0.75 & 14.40 & 14.44 & -0.04 &  0.51 \\ 
    A4 & Mistral-7B &  DO & 17.67  & 0.13 & 1.16 & 0.75 & 12.08 & 12.44 & -0.36 & 0.48 \\
    A5 & GPT-3.5 & DO & 18.05 & 0.62 & 0.39 & \underline{0.95}  & 15.12 & 15.06 & 0.06 & 0.56 \\
    A6 & GPT-4 & DO & 19.26  & \underline{0.84} & 0.66 & \textbf{0.99} & 19.12 & 16.99 & 2.13 & 0.71 \\
    A7 & GPT-4o & DO & \textbf{19.41} & \textbf{0.94} & \textbf{0.09 }& \textbf{0.99} &  16.26 & 15.38 & 0.88 & \underline{0.72}\\
    \hline\hline
        \multicolumn{10}{l}{\bf Few-Shot LLMs w/ 3 Demonstrations}\\
        \hline
      A8  & GPT-4 & DO  & 19.22  & 0.77 & 0.69 & 0.98 & 18.77 & 17.70 & 1.07 & 0.68\\
    A9 & GPT-3.5 & DO & 19.05& 0.54 & 0.41 & 0.77 & 16.07 & 16.17 & -0.10 & 0.54 \\

    \hline
    \multicolumn{10}{l}{\bf LLMs w/ Parameter-Efficient Fine-Tuning}\\
    \hline
    B1 & Flan-T5-XL & ED &\underline{ 22.22 }&  0.49 & 0.55 & 0.78 & 13.11 & \textbf{9.37} & \underline{3.59} & 0.70 \\

    B2 & Flan-T5-XXL  & ED & \textbf{23.43} & \underline{0.78} & \underline{0.27} & 0.85  & 14.40  & 10.78 & \textbf{3.62} & \textbf{0.75}  \\
    B3 & LLaMA2-7B & DO & 17.75 & 0.20 & 1.44 & 0.77 & 14.12 & 13.80 & 0.32 & 0.61\\
    B4 & Mistral-7B & DO & 18.21 & 0.29 & 1.76 & 0.83 & 13.10  & 12.42 & 0.68 & 0.55  \\
    \hline\hline
    \multicolumn{10}{l}{\bf Fully Fine-Tuned LLMs}\\
    \hline
    C1 & Flan-T5-Large-HP & ED & 21.36 & 0.56 & 0.53 & 0.86  & 13.71 & \underline{10.40} & 3.31 & 0.63 \\
    C2 & Flan-T5-Large-SP &  ED& 20.58 & 0.63 & 0.52 & 0.74 & 13.47 & 11.37 & 2.10 & 0.57 \\
    \hline\hline
    \multicolumn{10}{l}{\bf Two-Stage Modular Methods}\\
    \hline
    D1 & Flan-T5-XL-LoraHub & ED &  14.31 & -0.05 & 2.59 & 0.55 & 13.69 & 13.92 & -0.23 & 0.33\\
    D2 & Flan-T5-XL-AdapterFusion & ED & 17.65 & 0.04 & 1.40 & 0.65  & 14.31 & 14.49 & 0.18 & 0.41\\
    \hline

    \end{tabular}
    \label{tab:testcontrol}
\end{table*}

\section{Experiments}
\label{sec:results}

In this section, we benchmark and compare the compositional controllability of LLMs on  \benchname, focusing on four control attributes, which are categorized into explicit control (length, keywords) and implicit control (focus, readability). \textbf{\textit{As we focus on evaluating LLM's capabilities for compositional control, all experiments are conducted on the Compositional dataset ($\mathcal{D}_{CC}$) of \benchname, unless otherwise specified.}}

\paragraph{Implicit vs. Explicit Control}
As shown in Table~\ref{tab:testcontrol}, implicit control proves more challenging than explicit control. 

Most models struggle with implicit tasks. For readability, the $\delta_{FKGL}$ score remains below 1 across most models, indicating poor differentiation between readability levels. While GPT-4 achieves the highest $\delta_F$ of 2.13, its high FKGL score ($\sim$17) suggests it generates overly complex summaries, failing to improve readability as intended. Focus control is similarly weak, with most models scoring around 0.5 F1, and even GPT-4o, the best performer, reaching only 0.72 F1, highlighting limited adaptability.

In contrast, closed-source models excel at explicit control. GPT-3.5, GPT-4, and GPT-4o achieve strong length and keyword control, with PCC scores up to 0.94, MAD scores as low as 0.09, and success rates consistently exceeding 95\%. While open-source models underperform in length control, LLaMA2-7B and Mistral-7B still demonstrate strong keyword control.

\paragraph{Effect of Few-Shot Demonstrations}
Comparing zero-shot and few-shot performance of GPT-3.5 and GPT-4, we observe that few-shot demonstrations do not improve LLM capabilities in this task. This is likely due to the challenges of processing long sequences, where the length of scientific documents limits the number of effective in-context examples. As a result, the context window becomes constrained, and overly lengthy prompts reduce model effectiveness.

\paragraph{Effect of PEFT}

LoRA fine-tuning improves performance across all models, though decoder-only (DO) models show only minor gains in controllability. For example, $\delta_{FKGL}$ values for settings B3 and B4 remain below 1, indicating minimal differentiation in readability. Similarly, their length control remains weak, with PCC values below 0.3, reflecting a low correlation between generated and target lengths.

In contrast, encoder-decoder (ED) models B1 and B2 achieve results comparable to or surpassing GPT-4, despite weaker zero-shot performance. B2, in particular, exhibits strong control over length and keywords while significantly improving readability and focus—both implicit attributes. Under high readability constraints, its FKGL score drops below 11, aligning with a high school reading level. For focus control, while its F1 score of 0.75 leaves room for improvement, it still outperforms GPT-4.

\begin{figure}[t]
  \centering
  \includegraphics[width=7.6cm]{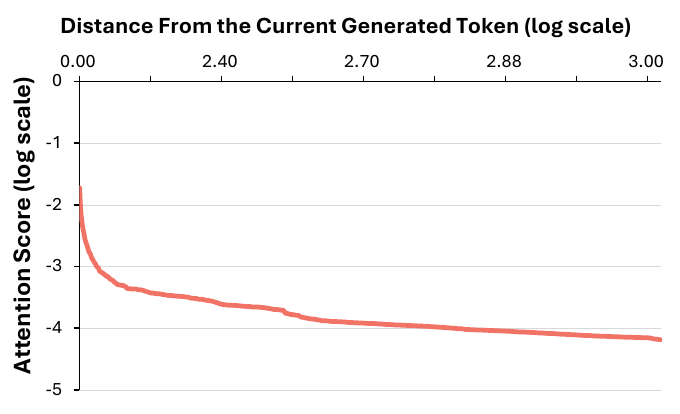}
      \caption{Average attention across all layers and heads for the last 10 tokens generated by LLaMA2 over the preceding 1,024 tokens, displayed on a log-log scale. The sharp decline in attention scores as token distance increases highlights the model's difficulty in maintaining focus on distant tokens within the sequence.}
  \label{fig:llama}
\end{figure}

\paragraph{Limitations of Decoder-Only Models in Compositional Controllable Summarization}
Our comparison of encoder-decoder models (B1, B2) and decoder-only models (B3, B4) shows that LLaMA2 and Mistral underperform compared to Flan-T5 under fine-tuning. This disparity likely arises from architectural differences. Encoder-decoder models like Flan-T5 are optimized for sequence-to-sequence tasks, where the encoder processes input while the decoder generates output by attending to the encoded representations. In contrast, decoder-only models such as LLaMA2 and Mistral handle both source and target sequences within the same unidirectional cross-attention mechanism~\cite{fu_2023_encoder_decoder}. As sequence length increases, attention becomes more diffused, weakening the model's ability to capture long-range dependencies—critical for controllable summarization.

To validate this, we analyze LLaMA2's attention patterns. Figure~\ref{fig:llama} shows a sharp decline in attention as token distance increases, indicating that the model prioritizes recent context while progressively disregarding earlier tokens. This degradation in long-range attention reduces the model’s ability to maintain focus on control signals and relevant input, ultimately impairing summarization quality.

\section{Discussion}
\label{sec:analysis}
\paragraph{Comparison of Single-Attribute vs. Compositional Control}
The results in Section~\ref{sec:results} reveal the limitations of LLMs in compositional control. To explore these limitations further, we compare the performance of Flan-T5-XXL fine-tuned on single-attribute tasks (i.e., fine-tuned separately on each of the four single-attribute datasets in \benchname) with its performance on compositional control (i.e., fine-tuned on $\mathcal{D}_{CC}$ as outlined in Section~\ref{sec:results}).

As shown in Table~\ref{tab:single}, performance drops significantly when multiple attributes are controlled simultaneously, suggesting a conflict between the attributes that limits the model's compositional controllability.

\begin{table}[t] 
    \centering
    \caption{Performance comparison of Flan-T5-XXL with LoRA on single and compositional control tasks.}
    \resizebox{\columnwidth}{!}{
    \begin{tabular}{lccccc}
    \hline
    \multirow{2}*{\bf Model} &  \bf Readability & \multicolumn{2}{c}{\bf Length} & {\bf Keywords} & {\bf Focus} \\
 \cmidrule(r){2-2} \cmidrule(r){3-4} \cmidrule(r){5-5} \cmidrule(r){6-6}
      & {$\delta_{F}$ $\uparrow$} & {PCC $\uparrow$} & {MAD $\downarrow$} & {SR $\uparrow$} & {F1 $\uparrow$} \\
    \hline
    Single  & 4.66 & 0.89 & 0.23 & 0.88 & 0.92 \\
    Compositional &  3.62 & 0.78 & 0.27 & 0.85 & 0.75  \\
    \hline
    \end{tabular}}

    \label{tab:single}
\end{table}

\begin{table*}[htbp]\small
    \centering
    \caption{Case studies on conflicts between readability and focus controllability. The left column presents unchanged attribute values, while the right column shows the impact of modifying a specific attribute. \textcolor{blue}{Blue} text highlights content that satisfies the control conditions, while \textcolor{red}{red} text indicates control failures.}
    \scalebox{1}{
    \begin{tabular}{p{4.5cm}|p{12.5cm}}
    \hline
    \multicolumn{2}{l}{Case 1: Length Against Readability -- Shorter Length Results in Unsatisfying Readability} \\
    \hline\hline
    \multirow{8}{4cm}{\textbf{Readability: high}\\Empirical Focus Level: low}  &
    {\textbf{Length Bin: 1} - We are interested in how the human visual system recognizes objects. \textcolor{blue}{Inspired by this, we created a new set of features that include information about gradients in both horizontal and vertical directions, as well as multi-scale contrast maps.}  These features are designed to help identify pedestrians in complex scenes.}\\
    \cline{2-2}
     & \textbf{Length Bin: 0} -
    We introduce a new set of features inspired by the human visual system, including multi-direction, multi-scale contrast and \textcolor{red}{statistical multi-channel descriptors}. These features are designed to enhance pedestrian detection performance in various conditions.\\
    \hline 
            \multicolumn{2}{l}{} \\
    \hline
    \multicolumn{2}{l}{Case 2: Readability Affects Focus -- Lower Readability Results in Less Empirical Focus}  \\
    \hline
    \hline
    \multirow{8}{4cm}{Length Bin: 0 \\ \textbf{Empirical Focus Level: high}} & 
    \textbf{Readability: high} - The effectiveness of the proposed human detection method \textcolor{blue}{was tested on several images. The results showed that the method was effective} even when the human's face was not visible, when the head was partially obscured, or when the subject was far from the camera.\\
    \cline{2-2}
     &  
    \textbf{Readability: normal} - This paper \textcolor{red}{presents a novel method} for detecting human presence in surveillance videos by analyzing the  \textcolor{red}{OMEGA symbol}. The proposed method involves identifying the unique signature of the human head-neck-shoulder region and classifying images as human or non-human based on this signature.\\
    \hline
        \multicolumn{2}{l}{} \\
    \hline
    \multicolumn{2}{l}{Case 3: Focus Affects Readability -- Lower  Empirical Focus Results in Lower Readability} \\
    \hline\hline
\multirow{7}{4cm}{Length Bin: 0 \\ \textbf{Readability: high}}            &\textbf{Empirical Focus Level: high} - 
    We applied our model to a large set of product reviews and \textcolor{blue}{achieved an accuracy of 98.28\%}, which is comparable to the current state of the art in sentiment classification.\\
        \cline{2-2}

    &\textbf{Empirical Focus Level:  low} - 
    This paper proposes a \textcolor{red}{supervised} sentiment classification model based on the \textcolor{red}{Naive Bayes algorithm}. It argues that a simple model can achieve comparable accuracy to state-of-the-art methods, highlighting the importance of selecting appropriate features and eliminating noise.\\

    \hline
    \end{tabular}}

    \label{tab:case 1}
\end{table*}

\begin{figure}[htbp]
  \centering
  \includegraphics[width=7.7cm]{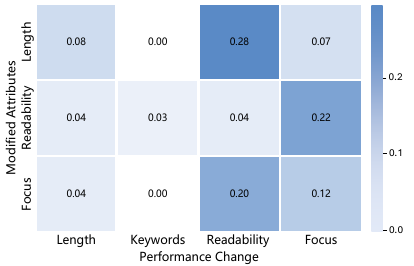}
      \caption{Attribute dependencies reflected by change amplitude (CA) on PCC (length), SR (keywords), $\delta_{FKGL}$ (readability), and F1 (focus). Each row represents the attribute being modified, and each column represents the affected attribute. Numerical values and color intensity denote the strength of the dependency. 
      }
  \label{fig:dependency}
\end{figure}

\paragraph{Bias in LLMs to Prioritize Explicit Attributes}
To better understand inter-attribute conflicts, we randomly select 300 samples from the test set of $\mathcal{D}_{CC}$ for each attribute. Keeping the other attribute settings unchanged, we modified the specified attribute (e.g., changing Readability from ``normal'' to ``high'') to create 300 new samples for each attribute accordingly. Then, we use the same model to generate the summary. For a given controllability metric, we calculate the change amplitude (CA) between the control ability before the attribute value change $m$ and after the change $m_{new}$, defined as:
$$CA=|\frac{m_{new} - m }{m}|.$$ 
We illustrate the dependencies between attributes in Figure~\ref{fig:dependency}, where each row represents the attribute being changed and each column represents the attribute being affected.  We observe that implicit attributes are more susceptible to the influence of other attributes compared to explicit attributes.

Specifically, in the case of Flan-T5, readability shows a stronger dependency on length than vice versa. This indicates that when controlling both attributes simultaneously, the model tends to prioritize the length constraint. For example, in Case 1 of Table~\ref{tab:case 1}, to shorten the length of the summary, the model omits the original simple explanation of the term ``feature'' and replaces it with the more abstract concept of ``descriptor'', making the summary more concise but harder to understand.

While implicit control requires more abstract reasoning, explicit control involves clearer and more concrete constraints. This bias suggests a limitation in the model's ability to deeply understand and reason across multiple attributes.

\paragraph{Conflicts Between Readability and Focus}
We observe a strong interaction between focus and readability, indicating that the model struggles to balance these two attributes. For example, in Case 2 of  Table~\ref{tab:case 1}, when the summary emphasizes empirical content, it tends to use simpler sentence structures and phrases such as \textit{``The results show...''} or \textit{``something was tested..''} along with terms like \textit{``effective''}. However, when readability requirements are reduced, the model introduces more complex, technical vocabulary, shifting the focus toward conceptual content.

In practice, when controlling both readability and focus, we expect the model to prioritize identifying content that aligns with the specified focus, rather than introducing incorrect content just to incorporate more technical language. Similarly, in Case 3, when a lower empirical focus level is set, the model uses more conceptual terminology, making the summary harder for a lay audience to understand. This suggests that the model has difficulty maintaining simple language while describing content with a low empirical focus.

\begin{figure}[t]
  \centering
  \includegraphics[width=0.95\columnwidth]{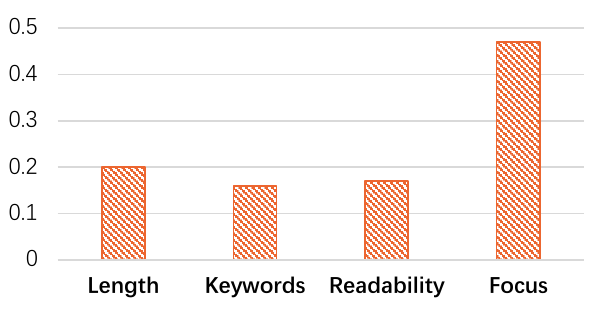}
      \caption{Average attention scores of the fusion layer for each adapter show a bias toward the focus adapter.}
  \label{fig:fusion}
\end{figure}

\paragraph{Can Two-Stage Modular Methods Address Inter-Attribute Conflicts?}

To address these trade-offs, we explore two-stage approaches where single-attribute control is learned in the first stage, and balancing multiple attributes is fine-tuned in the second stage. Specifically, we implement LoRAHub~\cite{huang2024lorahubefficientcrosstaskgeneralization}, which assigns a fixed weight to each LoRA module; and  AdapterFusion~\cite{pfeiffer-etal-2021-adapterfusion}, which utilizes an attention layer to learn how to focus on different attribute modules. However, results in Table~\ref{tab:testcontrol} show that neither AdapterFusion nor LoRAHub achieves meaningful controllability, with both producing poor summaries, as reflected in their low ROUGE scores. We believe this failure stems from the overly simplistic integration mechanisms, which are insufficient for learning the complexities required for compositional control, resulting in incoherent outputs.

We further probe the limitations of AdapterFusion by analyzing the model's attention differences across various attributes in  over the entire test set, as shown in Figure~\ref{fig:fusion}. The model exhibits a bias toward the focus module, suggesting that the attention mechanism struggles to adjust its emphasis on different attributes based on varying control requirements. This interference between modules ultimately degrades the model’s overall performance.

\begin{table*}[ht!] 
    \centering
    \caption{Comparison of GPT-4o’s performance on the first 200 samples of \benchname. \textbf{I+C}: using introduction and conclusion sections as input; \textbf{Full}: using the retrieved full texts as input. Detailed descriptions of metrics are provided in Section \ref{sec:eval_metrics}.}
    \begin{tabular}{llccccccccc}
    \hline
    & \multirow{3}*{\bf Method} & \multirow{3}*{\bf Quality} & \multicolumn{3}{c}{\bf Explicit Attributes} & \multicolumn{4}{c}{\textbf{Implicit Attributes}} & \multirow{3}*{\bf \#Tokens} \\
    \cmidrule(r){4-6} \cmidrule(r){7-10}
    &   &  &  \multicolumn{2}{c}{\bf Length} & {\bf Keywords} & \multicolumn{3}{c}{\bf Readability} & {\bf Focus} \\
    \cmidrule(r){3-3} \cmidrule(r){4-5} \cmidrule(r){6-6} \cmidrule(r){7-9} \cmidrule(r){10-10}
     & & {ROUGE-L $\uparrow$} & {PCC $\uparrow$} & {MAD $\downarrow$} & {SR $\uparrow$} & {F$_{normal}$} & {F$_{high}$$\downarrow$} & {$\delta_{FKGL}$ $\uparrow$} & {F1 $\uparrow$} &  \\
    \hline
    \hline
     & Full & \textbf{21.15}  & \textbf{0.98} & \textbf{0.06 }& 0.98 & 16.55 & 16.10 & 0.45 &  0.67 & 6.78k \\
    & I+C (ours) & 20.84 & 0.96 & \textbf{0.06} & \textbf{0.99} & 15.87 & \textbf{15.37 }& \textbf{0.50} & \textbf{0.72} & \textbf{1.61k} \\
    \hline
    
    \end{tabular}
    \label{tab:fulltext}
\end{table*}

\section{Discussion on Input Choice for Compositional Controllable Scientific Document Summarization}
\label{sec:input}

Obtaining high-quality summaries is key to effective scientific document summarization. To this end, we investigate the effects of using only the introduction and conclusion sections (I+C) versus the full scientific document as input for compositional controllable scientific document summarization.

\benchname adopts the I+C sections as input, a choice supported by prior research demonstrating its effectiveness over full-document summarization \cite{cachola-etal-2020-tldr, meng-etal-2021-bringing-facet}. This effectiveness likely stems from the challenges language models face in maintaining coherence and relevance over long contexts. As \benchname aims to foster effective general scientific summarization rather than produce in-depth analyses, the I+C sections provide a concise yet comprehensive representation of a paper’s key elements, including task definitions, prior limitations, methodologies, and main findings~\cite{stanfordtips}. Thus, our \benchname aligns with the established I+C practice in the field.

With recent LLMs such as GPT-4o \cite{hurst2024gpt} supporting extended context windows of up to 128k tokens, an open question arises: \textbf{Can these models overcome long-context challenges in processing full scientific documents for compositional controllable summarization?} To explore this, we conduct experiments on the first 200 samples of \benchname, retrieving their full documents and comparing GPT-4o’s summarization performance using introduction and conclusion sections (\textbf{I+C}) versus full-text input (\textbf{Full}).

The results, presented in Table~\ref{tab:fulltext}, indicate that \textbf{I+C} outperforms \textbf{Full} in keyword preservation, readability, and focus while maintaining comparable performance on other metrics. Notably, \textbf{I+C attains comparable performance while consuming only 23.7\% of the tokens required for Full, validating its effectiveness and practicality for compositional controllable scientific document summarization.}

\section{Conclusion}
We introduce \benchname, the first benchmark for evaluating compositional controllable scientific summarization. Integrating both explicit (length, keyword inclusion) and implicit (readability, empirical focus) attributes, \benchname provides a structured framework for assessing LLMs' ability to generate controlled, contextually appropriate scientific summaries. Our extensive experiments establish concrete LLM baselines and reveal significant limitations in compositional controllability, emphasizing the need for further research. We identify key challenges, including trade-offs between attributes and difficulties in handling implicit control, while also outlining potential directions for improvement. By providing actionable insights into LLMs' strengths and limitations in this challenging task, we offer a foundation for advancing the field.

\section*{Limitations and Future Work}
Empirical results on \benchname highlight key challenges in LLMs' ability to handle compositional controllable scientific summarization. While our findings show that fine-tuning smaller models (e.g., Flan-T5) can achieve performance comparable to GPT-4 across multiple metrics and provide deeper insights into failure cases, further research is needed to better understand and resolve observed interactions between control attributes.

Currently, \benchname focuses on scientific documents in English. Expanding this framework to support multilingual scientific summarization is a promising direction for future work. Additionally, while \benchname is designed for effective general communication, more granular summarization -- addressing the need for in-depth analysis, such as detailed experimental setups and side observations  -- remains an important area for exploration. Specifically, extending our framework and methodology to full scientific documents using datasets like LimGen~\cite{faizullah2024limgenprobingllmsgenerating} could provide valuable insights into more comprehensive and nuanced summarization strategies.

\section*{Acknowledgments}
We are grateful to the reviewers for their constructive feedback. This work was supported by the Singapore Ministry of Education Academic Research Fund Tier 1
(251RES2216).

\bibliographystyle{ACM-Reference-Format}
\bibliography{references}
\clearpage
\appendix

\section{Experimental Setup}
\label{app:experiments}
\subsection{Models}
Recent advancements in large-scale generative language models based on transformers, such as GPT~\cite{brown_2020_GPT}, have significantly improved performance on various natural language processing tasks. Given their widespread use and strong baseline performance, we select a range of large language models to evaluate their ability to perform compositional controlled summarization.

In our experiments, we focus on a variety of model architectures, including encoder-decoder models like the Flan-T5 series, decoder-only models such as LLaMA~\cite{touvron_2023_llama2openfoundation} and Mistral~\cite{jiang2023mistral7b}, and open-source large models like GPT-4~\cite{openai2024gpt4technicalreport}. Additionally, we evaluate their parameter-efficient fine-tuning (PEFT) versions to assess whether parameter-efficient fine-tuning enhances their ability to handle controlled summarization tasks. Moreover, we also revisit previously established baselines in the news domain for controlled summarization, namely hard prompt (HP, ~\citealp{zhang_2023_macsum}) and soft prefix tuning (SP, ~\citealp{li-liang-2021-prefix}) approaches on smaller models. We fine-tune each of these models using our compositional control dataset, specifically designed to test their ability to handle multiple attributes simultaneously. Across these models, we explore various objectives, yielding multiple model variants tailored for controlled summarization. These configurations allow us to comprehensively study the strengths and limitations of different architectures and tuning strategies when applied to compositional summarization tasks, as described in further detail below.

\paragraph{Zero-shot (ZS)}
In our simplest setting, we evaluate the compositional controllability already learned by large language models due to pretraining on large-scale corpora. In this setting, the models are not fine-tuned on any portion of our compositional control dataset. Instead, they must generate summaries for the test set based solely on the representations learned during pretraining.

At test time, the model receives the input document along with a set of control signals ($doc, len, focus, kw, read$) and generates the summary accordingly. The prompt template we used can be found in the Appendix~\ref{sec:prompt}.

\paragraph{Parameter-efficient Fine-tuning (PEFT)}
Because the pretraining domains of large language models are broad and cover diverse textual sources, we investigate whether adapting these models to the specific data distribution of our compositional control dataset can enhance their ability to handle multi-attribute summarization for scientific documents. In this setting, we apply parameter-efficient fine-tuning (PEFT) to these models,  which allows us to adapt these large models to our task without updating all of their parameters. In this setting, we leverage the widely-used PEFT technique LoRA (Low-Rank Adaptation, ~\citet{hu2021loralowrankadaptationlarge}), which inserts low-rank trainable matrices into the model's architecture. The document with the golden summary and four controlled attributes ($doc, summary, len, focus, kw, read$) is the input to the model.

\paragraph{Hard Prompt (HP)}
Following \citet{zhang_2023_macsum}, in the hard prompt setting, we provide the model with explicit instructions regarding the desired control attributes by using a structured prompt format. Each control attribute is formatted as "Attribute: Value", where “Attribute” refers to one of the specified control dimensions, such as "Length", "Focus", "Keywords", or "Readability", and “Value” represents the target value for that attribute (e.g., "Noramal", "High", etc.).

\paragraph{Soft Prefix (SP)}
In this setting, we prepend external trainable parameters-referred to as "prefixes"-to each layer of the summarization model to exert fine-grained control over the generation process. For controlling each attribute value, we assign $m$ prefix embeddings for the respective attribute, where $m$ is a hyperparameter representing the length of the prefix. For instance, for controlling summary readability with a "High" specification, we assign a series of embeddings $E_{Read:high} = [e^1_{Read:high}, ... , e^m_{Read:high}] $, where each $e_i^j$ is a vector of the same dimensionality as the model's word embeddings. For implementation details, readers may refer to ~\citet{li-liang-2021-prefix}.

\subsection{Implementation Details}
We use PyTorch and the Huggingface library to implement all the models. The experiments are conducted on 2 A40 GPUs. We choose 1e-4 as the learning rate for all PEFT models and the max epoch is set to 20. For hard prompt and soft prefix tuning, we largely follow the same training and inference setups as in \citet{zhang_2023_macsum}. In all of our experiments, we run each model once.

\subsection{Evaluation Metrics}
Here we introduce different metrics in detail.

\paragraph{Length Control}
For length controllability evaluation, we adopt (1)
the Mean of Absolute Deviation (MAD, ~\citealp{liu-etal-2018-length-ctrl-cnn}) of length codes of system-generated summaries and the references, measuring their length distance; and (2) the Pearson Correlation Coefficient (PCC, ~\citealp{liu-etal-2018-length-ctrl-cnn}) between the generated length and the input length bin. 

\paragraph{Readability Control} For evaluating readability controllabiltiy, we calculate the Flesch-Kincaid Grade Level (FKGL, ~\citealp{Kincaid1975DerivationON-fkgl}) for each document under both high and normal readability settings. A lower FKGL score indicate higher readability. We use $\delta_{F}$ to represent the difference in FKGL scores between the two categories. A larger $\delta_{FKGL}$ indicates a greater distinction in readability, which in turn reflects stronger control over readability by the model.
\paragraph{Keywords Control} For evaluating keyword controllability, we use the Success Rate (SR, ~\citealp{he2020ctrlsumgenericcontrollabletext}), namely the fraction of keywords actually occurring in the output summaries. We calculate SR employing exact matching after stemming.
\paragraph{Empirical Focus Level Control} 
Since focus is difficult to distinguish using simple rules, we opt to use an LLM evaluator for identification. For the generated summaries, we instruct GPT-4o to determine whether they are more conceptual-focused or empirical-focused. We experiment with various instruction patterns and validate the approach on the manually annotated dataset, achieving a best accuracy of 93\%. Next, we use the categories predicted by GPT-4o as the labels for the generated summaries and calculate the F1 score for each category. A higher F1 score indicates stronger control over the focus attribute. Prompt details can be found in Appendix~\ref{sec:prompt}.

\section{Human Evaluation Details}
\label{sec:human-eval}
As described in Section \ref{sec:ds-construction}, we employed human evaluators to assess the quality of GPT-generated summaries (regarding Readability and Focus) in \benchname.

The evaluations were conducted via Amazon Mechanical Turk (MTurk)\footnote{\url{https://www.mturk.com/}}, with participants compensated at a rate of USD \$0.60 per Readability case and USD \$0.75 per Focus case. We clearly stated during recruitment that the collected data would be used solely for research purposes. The human evaluation process has received IRB approval from the authors' institution.
\subsection{Instructions for Readability Evaluation}
\label{sec:read_eval}

\paragraph{Instructions to MTurk Annotators.}
Read the two scientific summaries and evaluate their facuality, readability and fluency.

      [Summary 1]
      
      [Summary 2]

\paragraph{Questions for MTurk Annotators.}

Q1: Do the two summaries agree on the main points?  (Yes / No)

Q2: Which summary is more readable?  (Summary 1 / Summary 2)

Q3: Is Summary 1 written in natural and fluent language?  (Yes / No)

Q4: Is Summary 2 written in natural and fluent language?  (Yes / No)

\subsection{Instructions for Focus Evaluation}
\label{sec:focus_eval}
\paragraph{Instructions to MTurk Annotators.}

Read the two scientific summaries and the reference abstract, evaluate their factuality, focus and fluency.

Definition:

The empirical-focused summary emphasizes the experimental, data-driven aspects of a study. It highlights data collection, experimental settings, performance metrics and concrete results obtained from the research, especially when it comes to statistics and data.  

The conceptual-focused summary emphasizes the theoretical and innovative aspects of a study. It highlights the motivation, the challenge, the underlying theoretical framework, the description of the proposed method or algorithm, and broader implications of the research.

Empirical-focused Example: This study evaluates the performance of a modified CRF-based POS tagging system for Manipuri, incorporating new features and the Reduplicated Multiword Expression (RMWE) feature. The experiment shows that the new CRF system achieves a Recall of 78.22\%, Precision of 73.15\%, and F-measure of 75.60\%. With the inclusion of RMWE as a feature, the results improve to a Recall of 80.20\%, Precision of 74.31\%, and F-measure of 77.14\%.

Conceptual-focused Example: This paper provides an in-depth overview of the updated feature selection approach in CRF for Manipuri POS tagging. It highlights the significance of optimal feature selection in enhancing CRF performance and discusses the introduction of new features, including the Reduplicated Multiword Expression (RMWE), which is crucial for accurately tagging Manipuri language POS due to its rich occurrence of RMWE.

      [Abstract]

      [Summary 1]
      
      [Summary 2]

\paragraph{Questions for MTurk Annotators.}

Q1: Does Summary 1 agree with the reference Abstract on the main points?  (Yes / No)

Q2: Does Summary 2 agree with the reference Abstract on the main points?  (Yes / No)

Q3: Which summary is more empirically-focused?  (Summary 1 / Summary 2)

Q4: Is Summary 1 written in natural and fluent language?  (Yes / No)

Q5: Is Summary 2 written in natural and fluent language?  (Yes / No)

\section{Prompt Templates}
\label{sec:prompt}

\subsection{Readability Data Construction}

"""

\textcolor{gray}{\# System}

You are a NLP expert. Please help me paraphrase some scientific summaries according my requests.\\ 

\textcolor{gray}{\# Instruction}

Please paraphrase this abstract for middle school students without using metaphors and without including information that cannot be obtained directly from the original abstract: \{abstract\}\\

"""

\subsection{Focus Data Construction}

"""

\textcolor{gray}{\# System}

You are a NLP expert. Please help me analyze some scientific abstracts.\\ 

\textcolor{gray}{\# Instruction for focus identification}

Can you identify the empirical content and conceptual content of the following abstract?  If the abstract is obviously predominated by only one type of information, the other part can be 'None'.\\

Definition:

The empirical content emphasizes the experimental, data-driven aspects of a study. It highlights data collection, experimental settings, performance metrics and concrete results obtained from the research, especially when it comes to statistics and data. 

The conceptual content emphasizes the theoretical and innovative aspects of a study. It highlights the underlying theoretical framework, the description of the proposed method or algorithm, and broader implications of the research.\\

Abstract: \{abstract\}\\

\textcolor{gray}{\# Output (Extracted by functions)}

\{"Empirical Content": output1, "Conceptual Content": output2\}\\

\textcolor{gray}{\# Instruction for paraphrasing}

If neither part is 'None', please paraphrase the empirical content and the conceptual content into a separated empirical-focused abstract and a separated conceptual-focused abstract respectively. Both abstracts should be coherent and fluent without including information that can't be inferred directly in the original abstract. If either part is 'None', take the input abstract and 'None' as the two output abstracts. Note that the difference between empirical focus and conceptual focus lies on the content rather than language expression. 
\\

\textcolor{gray}{\# Output (Extracted by functions)}

\{"Empirical Summary": output3, "Conceptual Summary": output4\}\\
"""

\subsection{Focus Evaluation}

"""

\textcolor{gray}{\# System}

You are a NLP expert. Please help me analyze some scientific abstracts.\\ 

\textcolor{gray}{\# Instruction}

Help me decide whether this abstract is more empirically-focused or more conceptually-focused. The output should be 'empirical' or 'conceptual'. \\
        
Definition:

The empirically-focused abstract emphasizes the experimental, data-driven aspects of a study. It highlights data collection, experimental settings, performance metrics and concrete results obtained from the research, especially when it comes to statistics and data. 

The conceptually-focused abstract emphasizes the theoretical and innovative aspects of a study. It highlights the motivation, the challenge, the underlying theoretical framework, the description of the proposed method or algorithm, and broader implications of the research.

Abstract: \{abstract\}\\

"""

\subsection{Zero-Shot LLM Experiments}
"""

\textcolor{gray}{\# System}

You are an expert summarizer who can generate summaries with specific controls.\\ 

\textcolor{gray}{\# Instruction}

Your task is to create a summary of the given scientific document with the following controls:\\

Length: The summary should fit in the specified word counts.

Keywords: Include the following keywords in the summary: [list of keywords]

Readability: Ensure the summary is either highly readable for laymen (high) or not specifically optimized for readability (normal).

Empirical Focus Level: Make the summary has high empirical focus level (high) by emphasizing the experimental, data-driven aspects of a study. It highlights data collection, experimental settings, performance metrics and concrete results obtained from the research, especially when it comes to statistics and data; or low empirical focus level (low)  the theoretical and innovative aspects of a study. It highlights the motivation, the challenge, the underlying theoretical framework, the description of the proposed method or algorithm, and broader implications of the research. \\

Document: \{doc\} \\

\textcolor{gray}{\# Control signals}

Length: 0-50 words

Keywords: "multi-scale"

Readability: high

Empirical Focus Level: high \\

\textcolor{gray}{\# Output}

Summary:

"""

\subsection{Few-Shot LLM Experiments}
"""

\textcolor{gray}{\# System}

You are an expert summarizer who can generate summaries with specific controls.\\ 

\textcolor{gray}{\# Instruction}

Your task is to create a summary of the given scientific document with the following controls:\\

Length: The summary should fit in the specified word counts.

Keywords: Include the following keywords in the summary: [list of keywords]

Readability: Ensure the summary is either highly readable for laymen (high) or not specifically optimized for readability (normal).

Empirical Focus Level: Make the summary has high empirical focus level (high) by emphasizing the experimental, data-driven aspects of a study. It highlights data collection, experimental settings, performance metrics and concrete results obtained from the research, especially when it comes to statistics and data; or low empirical focus level (low)  the theoretical and innovative aspects of a study. It highlights the motivation, the challenge, the underlying theoretical framework, the description of the proposed method or algorithm, and broader implications of the research. \\

\textcolor{gray}{\# Examples}

Examples: \\

Document: \{doc1\}

\{ctrl1\}

Summary: \{summary1\}\\

Document: \{doc2\}

\{ctrl2\}

Summary: \{summary2\}\\

Document: \{doc3\}

\{ctrl3\}

Summary: \{summary3\}\\

\textcolor{gray}{\# To generate}

Document: \{doc\} \\

Length: 0-50 words

Keywords: "multi-scale"

Readability: high

Empirical Focus Level: high \\

\textcolor{gray}{\# Output}

Summary:

"""

\end{document}